\documentclass{article}
\pdfoutput=1
     \PassOptionsToPackage{numbers, compress}{natbib}

\usepackage[utf8]{inputenc} 
\usepackage[T1]{fontenc}    
\usepackage{url}            
\usepackage{booktabs}       
\usepackage{amsfonts}       
\usepackage{nicefrac}       
\usepackage{microtype}      
\usepackage{natbib}
\usepackage[utf8]{inputenc}
\usepackage[english]{babel}
\usepackage[nonatbib, final]{neurips_2020}
 \usepackage{graphicx}
\graphicspath{ {./images/} }

\title{DeepWaste: Applying Deep Learning to Waste Classification for a Sustainable Planet}

\author{%
  \textbf{Yash Narayan} \\
  The Nueva School\\
  \texttt{yasnara@nuevaschool.org} \\
}

\begin{document}

\maketitle

\begin{abstract}
 
Accurate waste disposal, at the point of disposal, is crucial to fighting climate change. When materials that could be recycled or composted get diverted into landfills, they cause the emission of potent greenhouse gases such as methane. Current attempts to reduce erroneous waste disposal are expensive, inaccurate, and confusing. In this work, we propose \textbf{\emph{DeepWaste}}, an easy-to-use mobile app, that utilizes highly optimized deep learning techniques to provide users instantaneous waste classification into trash, recycling, and compost. We experiment with several convolution neural network architectures to detect and classify waste items. Our best model, a deep learning residual neural network with 50 layers, achieves an average precision of 0.881 on the test set. We demonstrate the performance and efficiency of our app on a set of real-world images. 

\end{abstract}

\section{Introduction}

Every year, the world generates over 2 billion tons of solid waste [1]. In the U.S., even though 75\% of this waste is capable of being recycled, only 34\% is actually recycled [2]. 
Further, 91\% of plastic isn't recycled [3] and only about 5\% of food and other organic waste is composted [4]. This waste generates over 1.5 billion metric tons of CO$_2$ equivalent greenhouse gases [1], contributing nearly as much to climate change as all the cars on the U.S. roads. 

Despite massive investment to educate the public about accurate waste disposal, efforts so far have been only moderately successful. People are often confused by what they can recycle, or compost. Signs and boards found ubiquitously near waste bins are difficult to understand and are often incomplete. Furthermore, disposal of waste varies based on the local recycling facilities' capabilities, and therefore rules for disposal are subject to change on a county-by-county basis. 

Errors in waste disposal constitute not only missed opportunities to recycle or compost, but also lead to the contamination of recycling and compost bins. Often, an entire bin can end up at a landfill due to a single error leading to contamination. Data from the National Waste and Recycling Association shows that human confusion in the identification and correct disposal of waste into our waste bins results in nearly 25\% of recyclables getting contaminated [5], diverting materials that could be recycled into our landfill. When a recyclable or compostable material ends up in the landfill, it releases methane which is several times more potent than CO$_2$ in contributing to global warming. Clearly, current ways to inform the public have not been working very well.  

In this work, we leverage the recent improvements of convolution neural networks (CNNs) for image-recognition tasks [6] and the availability of increased computational power on modern-day cell phones, to provide a novel approach for waste identification that is fast, low-cost, and accurate for anyone, anywhere. We present DeepWaste, the first mobile app targeted at the problem of erroneous waste disposal, at the point of disposal, through deep learning. We construct from scratch a fully-annotated dataset of more than 1200 waste items that is trained to achieve our best performing model with an average precision of 0.881 on the test set. 

\section{Previous Work}
\label{gen_inst}

The topic of waste classification has begun garnering some interest recently in research, but the attempts reported across literature to solve this problem have suffered from low accuracy (ranging from 22\% to low 70\%) [7], [8], or a network size that is too big for real-time application [9].  

Further, most of the previous attempts to mitigate the aforementioned problem of erroneous disposal envisaged deployment within a “smart bin” or the use of a commercial and industrial grade binning system within a recycling plant [10], [11], [12], requiring expensive hardware that costs thousands of dollars. The high cost of these solutions has so far been a deterrent to their large-scale adoption. Our approach is novel, as it allows the use of widely available mobile phones, and therefore has the potential of large-scale adoption at little or no cost.

Finally, none of the previous approaches have targeted compost classification. This is a significant problem because when compostable material such as food scraps and green waste gets into a landfill, it is generally compacted down and covered. This removes the oxygen and causes it to break down in an anaerobic process. Eventually, this releases methane, a greenhouse gas that is 25 times more potent than carbon dioxide in warming the earth over a 100-year timescale (and more than 80 times on a 20-year timescale) [13]. Our work is the first work in literature, to the best of our knowledge, that considers not only  recycling but also compost as a new category for classification.

\section{Methods}
\label{gen_inst}

Classifying waste using machine learning is challenging for three reasons.  First, whether a waste is recyclable or compostable or not depends on the properties of the material which can be hard to detect simply from the image. Second, the material can come in any shape such as a broken bottle, or a crumpled can, or deformed plastic; any machine learning technique needs to deal with this variation. Third, material that is recyclable depends on the capabilities of the local recycling center, so the app needs to take this geographical variable into account. 

Since there was no public dataset available, to accomplish this task, a dataset was constructed from scratch by contacting various recycling centers and collecting images from the local neighborhood. Towards this goal, we developed the ability for user generated content so that users can easily take a picture, annotate it, and upload it to the cloud for further training. In total, we manually collect 1218 images items at various lightings and angles, with 396 images containing compostable item(s), 427 images containing recyclable item(s), and 395 images containing landfill item(s). Utilizing this data, we experiment with several state-of-the-art convolution neural network methods, including InceptionV3 [14], Inception ResnetV2 [15], Resnet 50 [16], Mobile Net [17], and PNAS Net [18]. All CNNs used  were initialized with weights pre-trained on ImageNet. 

During training, each input image was rotated with an angle randomly selected among 0$^{\circ}$, 90$^{\circ}$, 180$^{\circ}$, and 270$^{\circ}$ and also randomly flipped, cropped, and blurred for data augmentation. Each method outputs a confidence for an inputted image. Hyper parameters specific to each method are set to the best values described in the original work.

\section{Results}
\label{headings}

Out of the various CNNs benchmarked on the dataset, Resnet50 showed the best accuracy and convergence on the test set in terms of average precision and thus was optimized and subsequently deployed inside of a mobile app using Apple CoreML.  Core ML optimizes on-device performance by leveraging the CPU, GPU, and Neural Engine while minimizing its memory footprint and power consumption. The DeepWaste model is running strictly on the user’s mobile device, therefore removing the need for internet connection and sharing data. 
The benchmarked models and their respective average precision scores on the test set can be found in Table 1.

\begin{table}[h!]
\begin{center}
\begin{tabular}{ |c|c|c|c|c|c| } 
 \hline
 \textbf{Accuracy} & \textbf{Inception\_V3} & \textbf{Inception\_ResnetV2} & \textbf{Resnet\_50} & \textbf{MobileNet} & \textbf{PNAS\_net} \\ 
 \hline
 Trash & 0.771 & 0.773 & 0.761 & 0.751 & 0.722 \\ 
 \hline
 Recycle & 0.891 & 0.783 & 0.924 & 0.949 & 0.864 \\ 
 \hline
 Compost & 0.806 & 0.806 & 0.882 & 0.873 & 0.841 \\
 \hline
 \hline
 \textbf{Overall} & \textbf{0.84} & \textbf{0.82} & \textbf{0.881} & \textbf{0.842} & \textbf{0.852} \\
 \hline
\end{tabular}
\end{center}
\caption{CNN performance on test set}
\label{table:1}
\end{table}

Figure 1 shows the DeepWaste app classifying commonly confused items in real-life. A user can simply point their phone camera to any piece of waste and get instantaneous feedback, with an average prediction time of around 100ms. DeepWaste is able to correctly identify items with high accuracy, even when the shape has been deformed such as a crushed soda can, orange peels, an apple core, crumpled paper, and a plastic bag. Note that the plastic bag in Figure 1 is classified as trash because plastic bags, films, and wraps cannot be recycled in your curbside recycling bin; they must be dropped-off to a special retail stores that can collect plastic grocery bags for recycling. Throwing this plastic bag into the recycling bin has the potential of contaminating the entire bin.

\begin{figure}
\begin{center}
  \includegraphics[width=13.2cm, height=5.7cm]{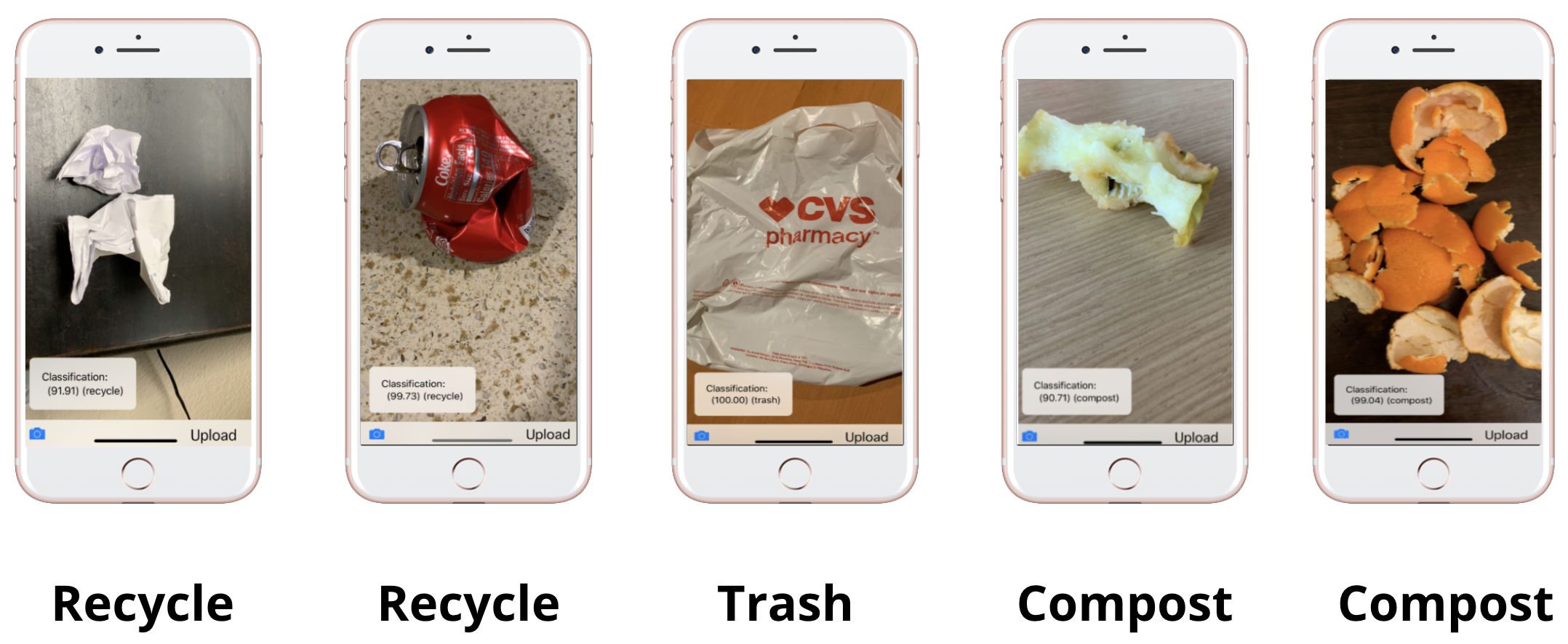}
  \caption{DeepWaste app classification output}
\end{center}
\end{figure}

\section{Conclusion and Future Work}
\label{others}
In this work, we propose a mobile application that uses highly-optimized deep learning techniques to provide users instantaneous waste classification, enabling them  to accurately dispose of waste into recycling, compost, or trash. Currently, the mobile app is available to beta users for testing. In the near future, we aim to construct a larger data set by releasing the app to the general public so that more users can add to the growing DeepWaste database. We have started a conversation with local recycling companies to explore if they would be interested in promoting the adoption of DeepWaste to their customers. The initial feedback is encouraging and we hope to continue this conversation and start a trial soon. We hope that our work can reduce the amount of incorrect waste disposal, and over time raise more awareness around the environmental impacts of waste on our climate. If DeepWaste can even reduce erroneous waste disposal by 1\%, it will be equivalent to removing over 6.5 million gasoline-burning passenger vehicles from the road, demonstrating the potential for machine learning techniques to tackle challenging problems related to climate change. 

\section{Acknowledgements}

I would like to thank Dr. Tanja Srebotnjak (The Nueva School), Professor Ram Rajagopal (Stanford University), Zhecheng Wang (Stanford University), Divyansh Jha (Esri), and Dr. Miranda Gorman (Project Drawdown) for their valuable feedback, guidance, and support.

\vspace{4.59mm}

\section*{References}

\medskip

\small

[1] S. Kazam, L. Yao, P. Bhada-Tata, F. Van Woerden, “What a Waste 2.0: A Global Snapshot of Solid Waste Management to 2050,” World Bank, pp. 3-5, 2018. 

[2] United States Environmental Protection Agency (EPA), “National Overview: Facts and Figures on Materials, Wastes and Recycling,” 2016. URL https://www.epa.gov/facts-and-figures-about-materials-waste-and-recycling/national-overview-facts-and-figures-materials

[3] L. Parker, “A Whopping 91 Percent of Plastic Isn’t Recycled,” National Geographic, 2018. URL https://www.nationalgeographic.org/article/whopping-91-percent-plastic-isnt-recycled/

[4] United States Environmental Protection Agency (EPA), “Food: Material-Specific Data,” 2016. URL https://www.epa.gov/facts-and-figures-about-materials-waste-and-recycling/food-material-specific-data

[5] M. Koerth, “The Era Of Easy Recycling May Be Coming To An End,” FiveThirtyEight, 2019. URL https://fivethirtyeight.com/features/the-era-of-easy-recycling-may-be-coming-to-an-end/

[6]  Y. LeCun, Y. Bengio, and G. Hinton, “Deep learning,” Nature, Vol. 521, pp. 436, 2015.

[7] G. Thung,  M. Yang, "Classification of Trash for Recyclability Status," \emph{Stanford CS229 Project Report}, 2016. 

[8] O. Awe, R. Mengitsu, V. Sreedhar, "Final Report: Smart Trash Net: Waste Localization and Classification," \emph{Stanford CS229 Project Report}, 2016.

[9] C. Bircanoğlu, M. Atay, F. Beşer, Ö. Genç and M. A. Kızrak, "RecycleNet: Intelligent Waste Sorting Using Deep Neural Networks," \emph{Innovations in Intelligent Systems and Applications (INISTA)}, Thessaloniki, pp. 1-7, doi: 10.1109/INISTA.2018.8466276, 2018. 

[10] I. Salimi, B. S. Bayu Dewantara, I. K. Wibowo, "Visual-based trash detection and classification system for smart trash bin robot," \emph{International Electronics Symposium on Knowledge Creation and Intelligent Computing (IES-KCIC)}, Bali, Indonesia, pp. 378-383, doi: 10.1109/KCIC.2018.8628499 2018. 

[11] D. Vinodha, J. Sangeetha, B. Cynthia Sherin, M. Renukadevi, "Smart Garbage System with Garbage Separation Using Object Detection," International Journal of Research in Engineering, Science and Management 2020.

[12] D. Ziouzios,  M. Dasygenis, "A Smart Recycling Bin for Waste Classification," \emph{Panhellenic Conference on Electronics \& Telecommunications (PACET)}, pp. 1-4, doi: 10.1109/PACET48583.2019.8956270, 2019.

[13] United States Environmental Protection Agency (EPA), "Overview of Greenhouse Gases," 2018. URL https://www.epa.gov/ghgemissions/overview-greenhouse-gases 

[14]  C. Szegedy,  V. Vanhoucke,  S. Ioffe, J. Shlens, Z. Wojna, "Rethinking the inception architecture for computer vision," in \emph{Proceedings of the IEEE conference on computer vision and pattern recognition}, pp. 2818-2826, 2016. 

[15] C. Szegedy, S. Ioffe, V. Vanhoucke, A.Alemi, "Inception-v4, inception-resnet and the impact of residual connections on learning," arXiv preprint arXiv:1602.07261, 2016. 

[16] K. He, X. Zhang, S. Ren, J. Sun, "Deep residual learning for image recognition," \emph{Proceedings of the IEEE conference on computer vision and pattern recognition}, pp. 770-778, 2016.

[17] A. Howard, M. Zhu, B. Chen, D. Kalenichenko, W. Wang, T. Weyand, M. Andreetto, Adam, H, "Mobilenets: Efficient convolutional neural networks for mobile vision applications," arXiv preprint arXiv:1704.04861, 2017. 

[18]  C. Liu, B. Zoph, J. Shlens, W. Hua, L. Li, L. Fei-Fei, A. Yuille, J. Huang, K. Murphy, "Progressive neural architecture search," arXiv preprint arXiv:1712.00559, 2017.

\end{document}